%% file: neurips_2026.tex
\documentclass{article}

\usepackage[dblblindworkshop, final]{neurips_2026}
\workshoptitle{Long-Context Foundation Models}

\usepackage[utf8]{inputenc} 
\usepackage[T1]{fontenc}    
\usepackage{hyperref}       
\usepackage{booktabs}       
\usepackage{amsfonts}       
\usepackage{nicefrac}       
\usepackage{microtype}      
\usepackage{xcolor}         

\usepackage{times}  
\usepackage{helvet}  
\usepackage{courier}  
\usepackage{graphicx} 
\usepackage{enumitem}

\usepackage{wrapfig}

\usepackage{natbib}  
\usepackage{caption} 
\usepackage{algorithm}
\usepackage{algorithmic}

\usepackage{graphicx}
\usepackage{subfigure}
\usepackage{booktabs} 
\usepackage{hyperref}
\usepackage{graphicx}
\usepackage{tikz}
\usepackage{bibentry}
\input{macros}

\usepackage{newfloat}
\usepackage{listings}
\DeclareCaptionStyle{ruled}{labelfont=normalfont,labelsep=colon,strut=off} 
\floatstyle{ruled}
\newfloat{listing}{tb}{lst}{}
\floatname{listing}{Listing}
\title{Efficient Hierarchical Transformers \\
for
Representing Log Data}

\author{
  Zhichao Hou\textsuperscript{\rm 1}\quad 
  Mina Ghashami\textsuperscript{\rm 2}\quad  
  Mikhail Kuznetsov\textsuperscript{\rm 2}\quad  \\
   \bfseries  Zifan Zhang\textsuperscript{\rm 1}\quad  
  MohamadAli Torkamani\textsuperscript{\rm 2} \\
    \and
  \textsuperscript{\rm 1}North Carolina State University, 
  \textsuperscript{\rm 2}Amazon Web Services\\
  \and
  \texttt{\{zhou4,zzhang66\}@ncsu.edu, 
  \{erata,lilienj,alitor\}@amazon.com}\\
}

\begin{document}

\maketitle

\input{section/abs}
\input{section/intro}

\input{section/method}

\input{section/exp}

\input{section/conclusion}


\newpage

\bibliographystyle{aaai25}
\bibliography{aaai25}


\appendix
\input{section/app}



\end{document}

%% file: macros.tex
\usepackage{mathtools}

\usepackage{amssymb}
\usepackage{amsthm}
\usepackage{color}

\makeatletter
\let\@@span\span
\def\sp@n{\@@span\omit\advance\@multicnt\m@ne}
\makeatother

\newcommand{\bc}{\begin{center}}
\newcommand{\ec}{\end{center}}

\newcommand{\bdm}{\begin{displaymath}}
\newcommand{\edm}{\end{displaymath}}

\newcommand{\beq}{\begin{equation}}
\newcommand{\eeq}{\end{equation}}

\newcommand{\bfl}{\begin{flushleft}}
\newcommand{\efl}{\end{flushleft}}

\newcommand{\bt}{\begin{tabbing}}
\newcommand{\et}{\end{tabbing}}

\newcommand{\beqn}{\begin{align}}
\newcommand{\eeqn}{\end{align}}

\newcommand{\beqs}{\begin{align*}} 
\newcommand{\eeqs}{\end{align*}}  



%% file: section/abs.tex
\begin{abstract}
Transformers have gained widespread acclaim for their versatility in handling diverse data structures, yet their application to log data remains underexplored. Log data, characterized by its hierarchical, dictionary-like structure, poses unique challenges when processed using conventional transformer models. Traditional methods often rely on manually crafted templates for parsing logs, a process that is labor-intensive and lacks generalizability. Additionally, the linear treatment of log sequences by standard transformers neglects the rich, nested relationships within log entries, leading to suboptimal representations and excessive memory usage.
To address these issues, we introduce \textit{HLogformer}, a novel hierarchical transformer framework specifically designed for log data. \textit{HLogformer} leverages the hierarchical structure of log entries to significantly reduce memory costs and enhance representation learning. Unlike traditional models that treat log data as flat sequences, our framework processes log entries in a manner that respects their inherent hierarchical organization. This approach ensures comprehensive encoding of both fine-grained details and broader contextual relationships.
Our contributions are threefold: First, \textit{HLogformer} is the first framework to design a dynamic hierarchical transformer tailored for dictionary-like log data. Second, it dramatically reduces memory costs associated with processing extensive log sequences. Third, comprehensive experiments demonstrate that \textit{HLogformer} more effectively encodes hierarchical contextual information, proving to be highly effective for downstream tasks such as synthetic anomaly detection and product recommendation.

\end{abstract}

%% file: section/intro.tex
\section{Introduction}
\vspace{-0.1in}
In recent years, transformers have garnered significant attention due to their versatility in handling various data structures, including images, text, graphs, tabular data, and temporal graphs~\citep{vaswani2017attention, dosovitskiy2020image, velivckovic2017graph,
huang2020tabtransformer, hou2026tefm, wu2024equivariant, hou2024protransformer} Despite their widespread application, there remains a notable gap in research focused on log data. Log data inherently possesses a hierarchical, dictionary-like structure, where each log entry is composed of nested fields and attributes. For instance, a single log entry might include metadata like timestamps, user IDs, and event types at the top level, while containing nested details such as specific actions taken, resources affected, and additional contextual information. Examples of log data include Amazon EC2 logs, IAM logs, and web server access logs.

Traditional methods for processing log data often involve manually applying templates to parse the logs before utilizing existing transformers. These templates are predefined rules or patterns designed to extract structured information from unstructured log messages. While this approach can be effective for certain types of logs, it has several limitations. Template-based methods can be labor-intensive, requiring domain-specific knowledge to create and maintain the templates. Additionally, they may not generalize well to diverse or evolving log formats, leading to incomplete or inaccurate parsing.

When lengthy log sequences are input into transformers for representation learning and downstream tasks~\citep{zhang2026centile}, several challenges arise. Firstly, the memory requirements become excessive due to the sheer volume of log data, making it difficult to process efficiently. Secondly, capturing the necessary contextual information demands larger and more complex transformer models, which can be computationally expensive and resource-intensive. Lastly, there is a tendency to treat log data as linear sequences, which neglects the hierarchical and structured nature of log entries. This linear treatment fails to leverage the rich, nested relationships inherent in log data, resulting in sub-optimal representation and analysis.

To address these challenges, researchers have proposed several approaches aimed at extending context length and reducing memory costs. Sparse transformers~\citep{child2019generating} leverage predefined patterns to limit the number of attention connections each token has. Local attention restricts the attention mechanism to a fixed-size window around each token, ensuring that only nearby tokens are considered. This approach is efficient for capturing local dependencies and reduces the overall computational burden. Strided attention extends this idea by allowing tokens to attend to other tokens at fixed intervals, further reducing the number of attention connections while maintaining the ability to capture broader context across the sequence. Other methods, such as the ones proposed by~\citep{roy2021efficient} and \cite{kitaev2020reformer}, take this concept further by making the sparsity pattern learnable.

Additionally, models like Longformer~\citep{beltagy2020longformer}, ETC (Extended Transformer Construction)~\citep{ainslie2020etc}, and Big Bird~\citep{zaheer2020big} introduce global memory tokens to address the limitations of traditional transformers in handling long sequences. These global memory tokens are specialized tokens that have attention connections to all other tokens in the sequence. This mechanism enables the models to maintain a broader contextual understanding without the quadratic memory and computational overhead typically associated with the self-attention mechanism in standard transformers.
There are techniques such as Transformer-XL~\citep{dai2019transformer} and Compressive Transformer~\citep{rae2019compressive} which employ segment-based recurrence to significantly reduce memory and computational costs. Despite their effectiveness, these approaches are not tailored to the unique characteristics of log data.

There are several hierarchical transformers~\citep{nawrot2021hierarchical, pappagari2019hierarchical, pan2021scalable, liu2021swin} that modify the vanilla transformer architecture to obtain hierarchical representations of the data. However, these architectures primarily build the hierarchy by encoding the tokens using downsampling, pooling, or segmentation techniques, which are not specifically designed for the hierarchical log data we are interested in.

In this paper, we introduce a novel and efficient hierarchical transformer framework specifically designed for log data, termed \textit{HLogformer}. Our \textit{HLogformer} framework addresses the unique challenges of log data by significantly reducing memory costs, making it feasible to apply transformers to lengthy log sequences. Furthermore, \textit{HLogformer} captures and leverages the inherent hierarchical structural information within the data, thereby enhancing representation learning. Our key contributions are:
\vspace{-0.05in}
\begin{itemize}[left=0.0em]
    \item 
  
    To the best of our knowledge, \textit{HLogformer} is among the first frameworks to build a hierarchical transformer that directly follows the key–value tree of dictionary-structured (e.g., JSON) logs.
    \item \textit{HLogformer} dramatically reduces memory costs associated with processing extensive log data.
    \item Comprehensive experiments demonstrate that \textit{HLogformer} more effectively encodes hierarchical contextual information, proving to be highly effective for downstream tasks such as synthetic anomaly detection and product recommendation.
\end{itemize}
\vspace{-0.05in}
The rest of the paper is organized as follows. Section 2 reviews the related work, providing context and background that underpins our study. In Section 3, we delve into the proposed methodology and training strategy, detailing the innovative approaches and techniques we employ. Finally, Section 4 presents the experiments and results, showcasing the effectiveness and practical implications of our proposed model.

%% file: section/method.tex
\vspace{-0.15in}
\section{Proposed Methodology}
In this section, we discuss the hierarchical structure inherent in log data and introduce our novel model, \textit{HLogformer}, designed to leverage this structure. 

\subsection{Hierarchical Structure of Log Data}

\begin{figure}[h!]
\centering
\vspace{-0.1in}
\subfigure[Log as a sequence]{
\includegraphics[width=0.48\textwidth]{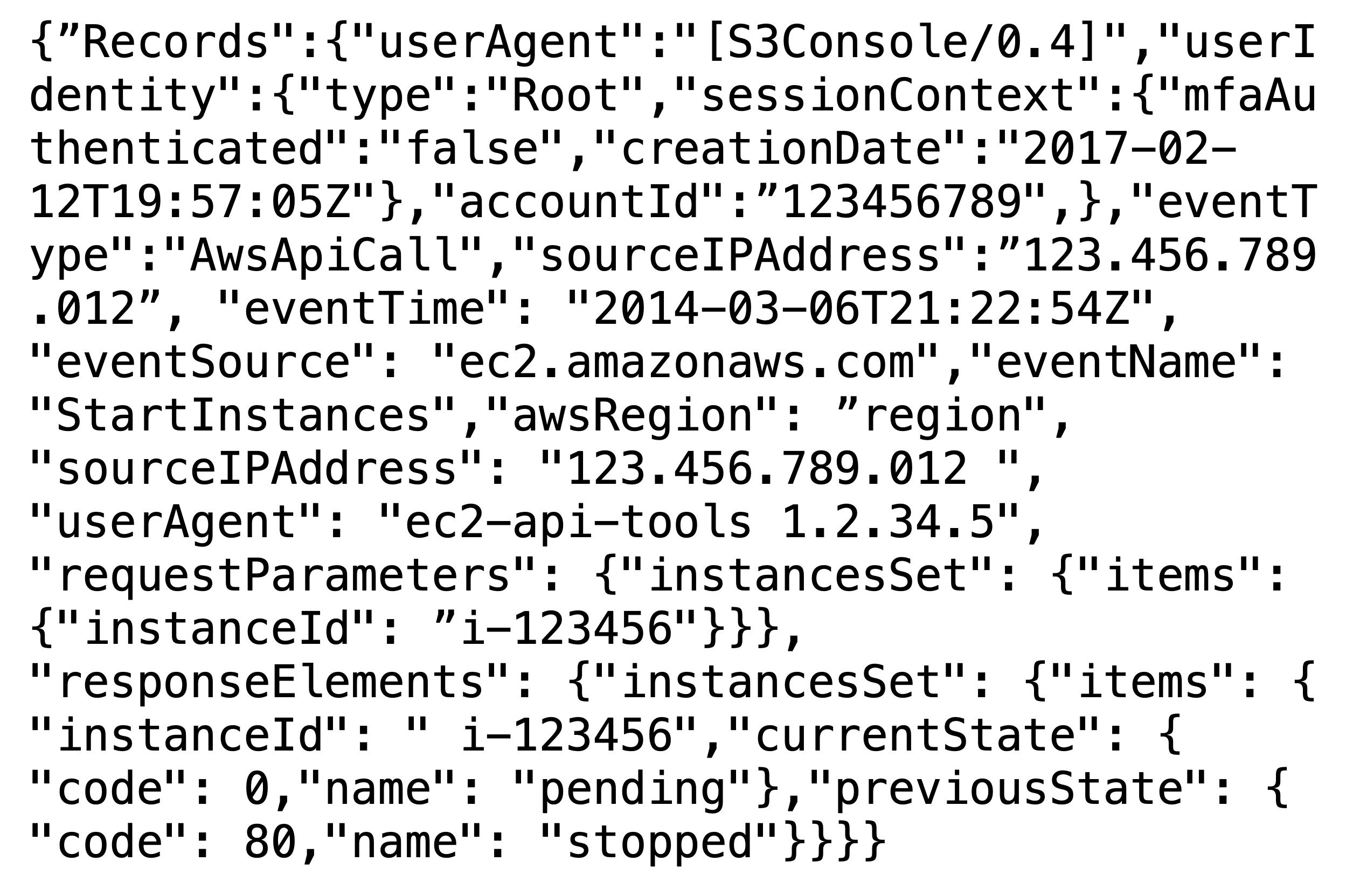}
}
\hfill
\subfigure[Log as a hierarchical tree]{
\includegraphics[width=0.48\textwidth]{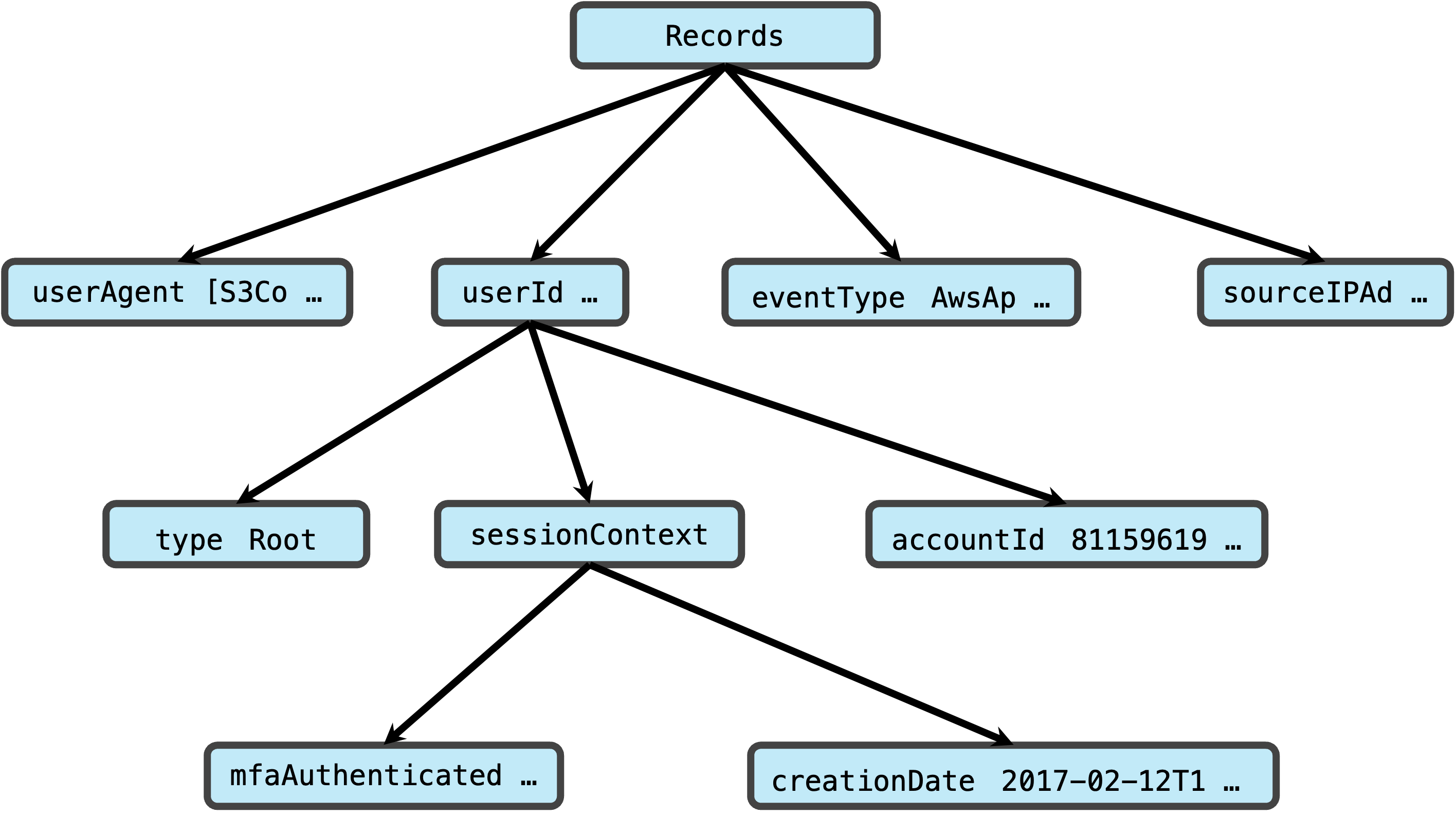}
}
\vspace{-0.1in}
\caption{Different representations of log data: (a) treating log data as a sequence, and (b) treating the log data as a hierarchical tree.}
\label{fig:diff_data_representation}
\vspace{-0.1in}
\end{figure}

As illustrated in Figure~\ref{fig:diff_data_representation}, log data such as AWS CloudTrail logs can be represented either as a linear sequence (Figure~\ref{fig:diff_data_representation}(a)) or a hierarchical tree (Figure~\ref{fig:diff_data_representation}(b)). While the sequential representation enables the use of conventional language models by treating log entries as tokens, it may obscure nested relationships among fields. For example, the \texttt{userIdentity} field contains attributes such as account ID, username, principal ID, and \texttt{sessionContext}, which itself contains nested attributes. In contrast, the tree representation preserves these parent--child relationships, enabling a richer and more accurate characterization of multi-level dependencies within log data.


\subsection{HLogformer: A Hierarchical Log Transformer} 
\begin{figure}[h!]
    \centering
    \includegraphics[width=0.9\linewidth]{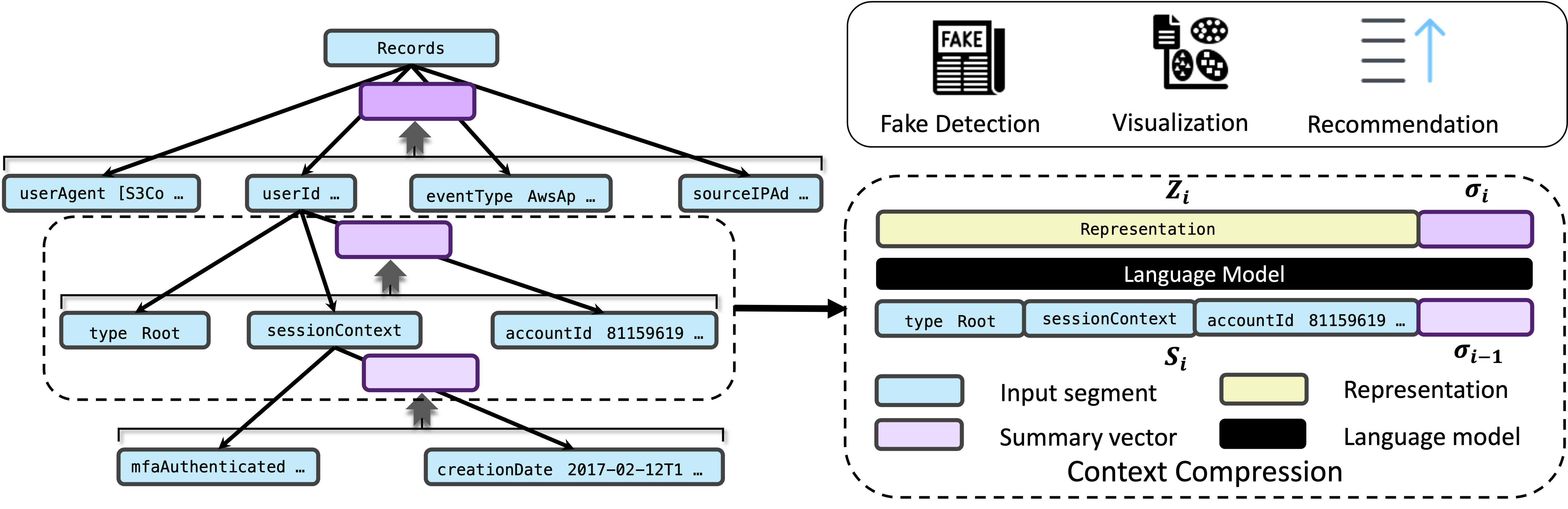}
    \caption{Schematic overview of HLogformer: HLogformer encapsulates the context segment into a summary vector, which is then passed from low-level to high-level (left). Specifically, at each step, we concatenate all the child nodes' tokens $S_i$ and the previous summary vector $\sigma_{i-1}$ as the input. The language model is then applied over this input to obtain the updated summary vector and the token representation (right). }
    \label{fig:hlogformer_arch}
\end{figure}
To fully leverage the hierarchical structure inherent in log data, we introduce a novel architecture called \textit{HLogformer}, illustrated in Figure~\ref{fig:hlogformer_arch}
. This architecture is inspired by context compression techniques~\citep{chevalier2023adapting}, but unlike them, \textit{HLogformer} segments log data according to its hierarchical tree structure. This segmentation process progresses systematically from low-level details to high-level summaries, mirroring the natural organization of the data. Each segment corresponds to a distinct level of the hierarchical structure, ensuring that the model respects and utilizes the nested relationships within the log entries.

We can first represent the log data as a directed graph $\mathcal{G}=(\mathcal{V},\mathcal{E})$ where $s_i$ denotes the text in node $v_i\in \mathcal{V} $ while $e_{ij}=(v_i,v_j)\in\mathcal{E}$ denotes the parent-child relationship in the log data. For step $i$, we concatenate all the child nodes' text of node $i$ as the segment $S_i=\textbf{Concat}[\{s_j: e_{ij}\in\mathcal{G}\}]$.

The processing pipeline of \textit{HLogformer} operates step-by-step as shown in Figure~\ref{fig:hlogformer_arch} (right), beginning with the most granular details of the log data. At each step, the architecture processes a segment of the log data, extracting and summarizing the relevant information. These summary vectors encapsulate the essential context and dependencies at the current level of the hierarchy. Once processed, these summary vectors are passed to the next step, where higher-level segments are processed similarly.
At each step $i$, the segment $S_i$ is processed along with the summary vector from the previous step $\sigma_{i-1}$. This process ensures that the hierarchical context is preserved and progressively refined as we move through the log data. The following equation formalizes this process, where the log data segment $S_i$ and the summary vector from the previous step $\sigma_{i-1}$are combined and processed by the language model \textbf{LM}: 
\begin{equation}
\label{eq:summarization}
    Z_i,\sigma_{i} = \textbf{LM}([S_i,\sigma_{i-1}])
\end{equation}
In this equation, \textbf{LM} represents the language model that generates the new summary vector $\sigma_i$ and the intermediate representation $Z_i$, capturing both the current segment's information and the accumulated context from previous segments. 

\textbf{Bidirectional Hierarchical Compression Paradigm.} 
In the primary architecture described above, summary vectors are passed exclusively from low-level to high-level segments. This allows high-level tokens to access low-level information through the summary vectors, but it may result in low-level tokens missing out some corresponding high-level context. 
To address this limitation, we propose a bidirectional summary passing technique. This involves initially passing the summary from low-level to high-level, and then reversing the process to ensure that low-level tokens can also benefit from high-level information.


\textbf{Complexity Analysis.} Our \textit{HLogformer} provides an efficient framework for handling long context in log data. Assume the entire sequence has a length of $L$ and is split into $M$ equal-sized segments. Then the vanilla transformer has a memory complexity of $O(L^2)$, while \textit{HLogformer} reduces this to $O(L^2/M)$.

\textbf{Advantages.}
This progressive approach offers several key advantages: (1) By segmenting the log data according to its hierarchical structure, \textit{HLogformer} captures both fine-grained details and broader contextual relationships, building a comprehensive and layered representation at each step; (2)
This method significantly reduces memory and computational costs by summarizing information at each level and passing only the accumulated summary vectors to the next step, efficiently managing the data's complexity and size;
(3) Additionally, \textit{HLogformer} enhances the model's ability to perform downstream tasks such as anomaly detection, log classification, and predictive maintenance. By maintaining and leveraging the hierarchical structure, the model can more accurately identify patterns and anomalies within the data.

\vspace{-0.1in}
\subsection{Training Strategy} 
\begin{figure}[h!]
    \centering
    \vspace{-0.1in}
    \includegraphics[width=0.7\linewidth]{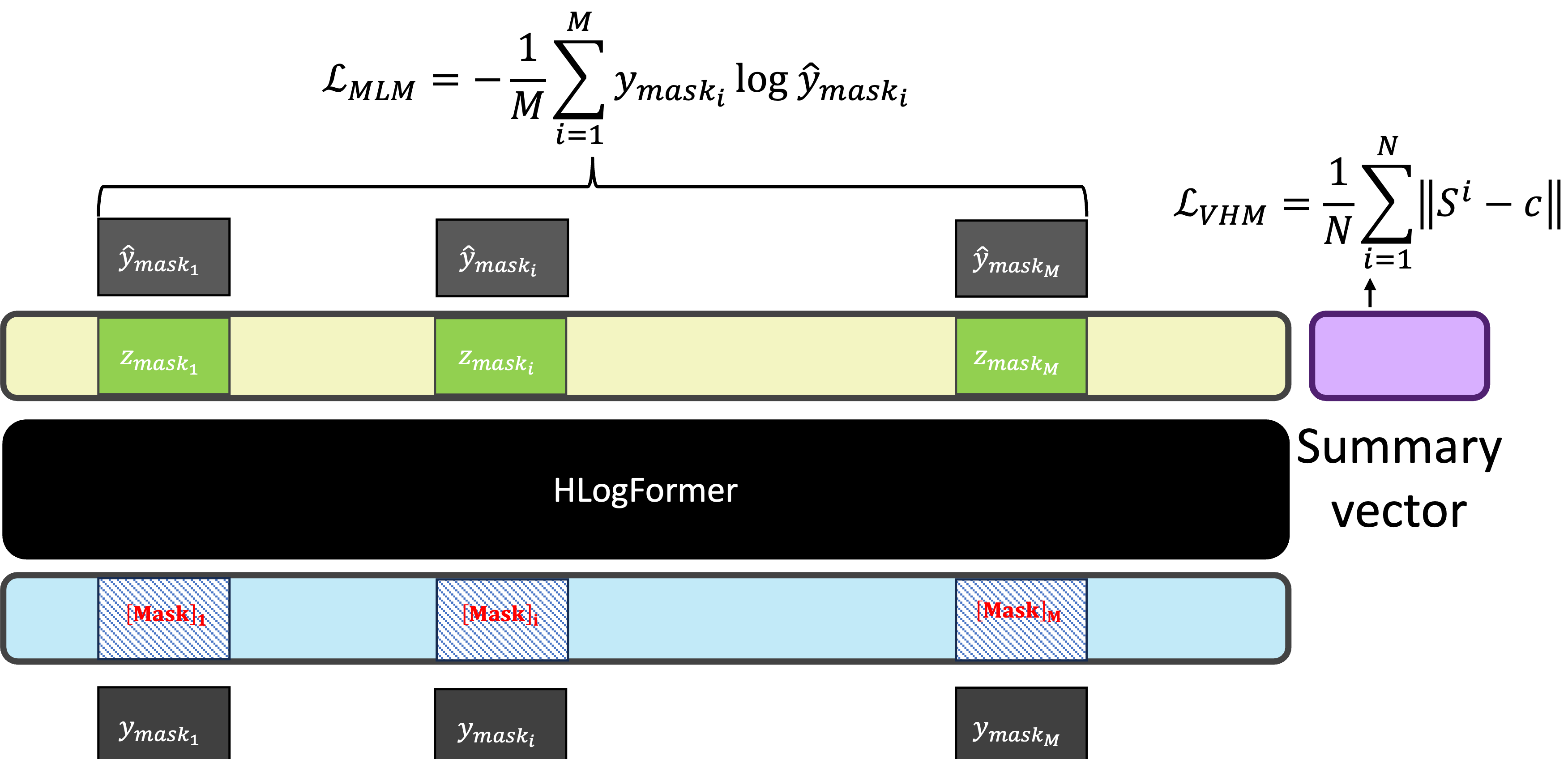}
    \caption{Self-supervised Learning.}
    \label{fig:self_supervised_learning}
    \vspace{-0.15in}
\end{figure}

After building the hierarchical log transformer, we need to adopt an appropriate training strategy to obtain informative representations and perform downstream tasks. Given that log data typically lack labels, we propose a \textit{self-supervised learning approach} using masked language modeling loss and volume hypersphere minimization loss, as illustrated in Figure~\ref{fig:self_supervised_learning}.

\textbf{Masked Language Modeling.}
To capture the contextual information of log data, we utilize the masked language modeling (MLM) task, which has proven effective in various natural language processing applications. This approach involves randomly selecting a subset of tokens from the input data and replacing them with a special \texttt{[MASK]} token. The model is then tasked with predicting the original tokens that were masked, allowing it to learn rich contextual representations of the log data.
The training objective for this task is defined by the cross-entropy loss function, which measures the discrepancy between the predicted tokens and the actual tokens at the masked positions. Formally, the MLM loss is expressed as:
$\mathcal{L}_{MLM}=\frac{1}{M}\sum_{i=1}^My_{mask_i}\log \hat{y}_{mask_i},$
where $M$ is the number of masked tokens, $y_{mask_i}$ represents the actual token at the $i$-th masked position, and $\hat{y}_{mask_i}$ is the predicted token at the same position. This loss function encourages the model to accurately predict the masked tokens, thereby forcing it to learn the underlying patterns and dependencies in the log.

\textbf{Volume Hypersphere Minimization.} 
Given our assumption that all training data represents real or normal instances, the task aligns well with one-class classification problems. In this context, we draw inspiration from the One-Class Deep SVDD~\citep{ruff2018deep} methodology. Our objective is to map normal data points as closely as possible to the center of a hypersphere. This approach effectively captures the notion of normality by ensuring that the representations of normal data points are densely clustered.
To achieve this, we seek to minimize the volume of the hypersphere by positioning its center, denoted as $c$, such that the mean distance of all data representations to this center is minimized. Formally, this minimization problem is expressed through the following loss function:
$\mathcal{L}_{VHM}=\frac{1}{N}\sum_{i=1}^N\|S^i-c\|,$
where $N$ is the number of data points, $S^i$ represents the accumulated summary vector of $i$-th data point, and $c=\frac{1}{N}\sum_{i=1}^N S^i$ is the calculated center of all the data representations. This center $c$ is dynamically computed as the average of all summary vectors, ensuring that it accurately reflects the central tendency of the normal data points.

By minimizing this loss, we encourage the model to produce representations that are not only compact but also concentrated around a central point. This is crucial for downstream tasks such as anomaly detection, where deviations from this central cluster can be effectively identified as anomalies.

%% file: section/exp.tex
\vspace{-0.1in}
\section{Experiments}
\vspace{-0.1in}

In this section, we will begin by evaluating the effectiveness of our \textit{HLogformer} model in masked language modeling tasks. This will involve assessing its ability to accurately predict masked tokens within a sequence, thereby demonstrating its understanding of the underlying log data. Following this, we will apply our model to several downstream tasks to further validate its utility and performance. These tasks include fake log detection, where the model will be tested on its capability to identify fraudulent or synthetic log entries, and visualization analysis, where we will leverage the model’s outputs to generate insightful visual representations of the log data.
\subsection{Experimental Setting}
In this section, we detail the dataset utilized for our experiments, the backbone architecture underpinning our models, and the hyperparameters selected to optimize performance.

\textbf{Datasets.} We use the following datasets in our experiments:
\noindent(1) \href{https://summitroute.com/blog/2020/10/09/public_dataset_of_cloudtrail_logs_from_flaws_cloud/}{CloudTrail Logs Dataset}: It is an anonymized public log data from \texttt{flaws.cloud} that covers over 3.5 years of data and 1,939,207 number of events. 
\noindent(2) OKTA: This log data is a private dataset which monitors and audits authentication activity to an internal system.
\noindent(3) \href{https://github.com/adanalvarez/TrailDiscover/tree/main}{TrailDiscover}: It is an evolving repository of CloudTrail events with detailed descriptions, MITRE ATT\&CK insights, real-world incidents, references and security implications. 
\noindent(4) \href{https://amazon-reviews-2023.github.io/main.html}{Amazon Reviews}~\citep{hou2024bridging}: It is collected in 2023 by McAuley Lab. We use 9 categories of Item Metadata in Amazon Reviews including All Beauty, Amazon Fashion, Appliances, Arts Crafts and Sewing, Automotive, CDs and Vinyl, Digital Music, Health and Personal Care, and Magazine Subscriptions.

\textbf{Backbone Architectures.} 
Since our \textit{HLogformer} is designed as a versatile plugin~\citep{hou2024protransformer, hou2025robustness} capable of integrating with any transformer architecture, we will experiment with a variety of backbone models to demonstrate its adaptability and effectiveness. Specifically, we will employ several transformer architectures, including the vanilla Transformer~\citep{devlin2018bert} with random initialization, pretrained Transformer~\citep{devlin2018bert} with pretrained parameters in \texttt{bert-base-uncase}    and four efficient transformers: Linear Transformer~\citep{katharopoulos2020transformers}, Reformer~\citep{kitaev2020reformer}, Routing Transformer~\citep{roy2021efficient}, and Sparse Transformer~\citep{child2019generating}.

\textbf{Hyperparameters.} \label{hyperparam}
 We use 8 transformer blocks for backbone models and 1 block for our \textit{HLogformer}. We set the number of training epochs to 100, the masking rate to 0.2, and the length of the summary vector to 10 tokens. We use Adam optimizer with the learning rate of \{0.01,0.005, 0.001 \}, the Adam weight decay of \{0.01, 0.001,0.0001\}, Adam $\beta_1$ of \{0.3,0.6, 0.9\}, and Adam $\beta_2$ of \{ 0.9, 0.99, 0.999\}. For each dataset, the training/validation/testing ratio is set as 5:1:1.

\vspace{-0.1in}
\subsection{Self-Supervised Learning Task}

\begin{table}[h!]
\vspace{-0.2in}
\caption{ Masked language modeling loss on security datasets. }
\centering
\resizebox{0.8\linewidth}{!}{
\begin{tabular}{l|c|c|c|c}
\toprule
\textbf{Architecture} & \textbf{CloudTrail} & \textbf{OKTA}&\textbf{TrailDiscover}& \textbf{\#Parameter } \\
\hline
Vanilla Transformer & 5.692 &4.221&5.676& 12636160 \\
Vanilla-\textit{HLogformer} (Ours)  & 4.158 &2.888&4.921& 789760 \\
\hline
Pretrained Transformer&4.414&3.872&5.078&12636160\\
Pretrained-\textit{HLogformer} (Ours)&3.850&2.611&4.995&789760 \\
\hline
Linear Transformer & 5.341 &4.449&5.786& 12636160 \\
Linear-\textit{HLogformer} (Ours) & 4.092 &2.833&5.101& 789760 \\
\hline
Reformer & 4.184  &3.504&5.460& 13602048 \\
Reformer-\textit{HLogformer} (Ours) & 4.106 &3.215&5.004& 8537856 \\
\hline
Routing Transformer & 6.937 &5.313&9.716& 10522624 \\
Routing-\textit{HLogformer} (Ours) & 4.186 &2.748&5.323& 1315328 \\
\hline
Sparse Transformer & 8.421  &5.446&8.871& 4212736 \\
Sparse-\textit{HLogformer} (Ours)& 4.766 &3.789&5.508& 526592 \\
\bottomrule
\end{tabular}
}

\label{table:mlm_loss}
\end{table}

\begin{table}[h!]
\centering
\vspace{-0.2in}
\caption{Masked language modeling loss on Amazon Review datasets. }

\begin{footnotesize}
    
\setlength{\tabcolsep}{4.5pt}
\begin{tabular}{l|c|c|c|c|c|c|c|c|c}

\toprule
&\textbf{Beauty}& \textbf{Fashion}& \textbf{Appliances}& \textbf{Arts} & \textbf{Auto}& \textbf{CDs}& \textbf{Music}&\textbf{ Health} & \textbf{Magazine} \\
\hline
Vanilla Transformer &4.571 & 4.82 & 5.029 & 5.267 & 5.018 & 4.345 & 4.316 & 5.021 & 4.054\\
Vanilla-\textit{HLogformer}  & 3.686 & 3.46 & 3.999 & 4.372 & 4.57 & 3.449 & 3.565 & 3.936 & 3.159\\
\hline
Linear Transformer & 4.690&	4.668	&5.078&	5.124&	6.176&4.208	&4.360	&5.023&4.126\\
Linear-\textit{HLogformer}  & 3.758	&3.871	&3.695	&4.285	&4.841&3.425&	3.499	&3.839&2.963\\
\hline
Reformer & 4.069 &	4.212	&4.389&	4.581&	4.741&	3.545&	3.978&	4.349&3.283 \\
Reformer-\textit{HLogformer}  & 3.593&	4.003&	4.014&	4.095	&4.150	&3.386&	3.540&	3.976	&2.785 \\
\hline
Routing Transformer & 8.494&8.503	&7.871&	8.771&8.665	&7.454&	7.532&	8.561&7.400\\
Routing-\textit{HLogformer}  & 3.773&3.955&	4.196&	4.473&4.470&	3.538	&3.521	&4.173&3.253 \\
\hline
Sparse Transformer & 9.680	&7.654&	7.470&	10.196&9.666&	8.629&	9.561&	9.567&8.687 \\
Sparse-\textit{HLogformer}  &4.127 &3.994	&5.071	&4.967&4.798&	3.718&	4.147&	4.611&3.837 \\
\bottomrule
\end{tabular}
\end{footnotesize}

\label{table:mlm_loss_amazon}
\end{table}

To demonstrate the effectiveness and efficiency of our HLogformer, we train the models with masked language modeling loss  and  present the loss and the number of parameters in the transformer block in Table~\ref{table:mlm_loss} (Security datasets) and Table~\ref{table:mlm_loss_amazon} (Amazon Reviews datasets). From these tables, we can make the following observations:
\begin{itemize}[left=0.0em]
\item Our hierarchical framework is a highly effective plug-in module that significantly and consistently reduces masked language modeling loss and therefore improves the ability to capture contextual information.
\item Our \textit{HLogformer} requires only a small-sized transformer block while achieving better results than the backbone models. As we mentioned in the hyperparameters section, we use only 1 transformer block in \textit{HLogFormer} to handle the segments at each step, while the backbone models require 8 blocks to be able to process the large log data.
\end{itemize}

\subsection{Supervised Learning Task on TrailDiscover}


\begin{wraptable}{r}{0.6\textwidth}
\vspace{-0.15in}
\caption{Average accuracy of supervised classification task. Task 1 is the binary classification task on \texttt{"usedInWild"} and Task 2 is the multi-class classification task on \texttt{"MITRE Attack Tactics" }.}
\centering

\resizebox{0.8\linewidth}{!}{
\begin{tabular}{l|c|c}
\toprule
\textbf{Architecture} & \textbf{Task 1}&\textbf{Task 2} \\
\hline
Vanilla Transformer &67.059&  69.412 \\
Vanilla-\textit{HLogformer} (Ours)  & 95.294& 77.647 \\
\hline
Linear Transformer &65.882&51.765  \\
Linear-\textit{HLogformer} (Ours)  &92.941&57.647  \\
\hline
Reformer &65.882&70.588 \\
Reformer-\textit{HLogformer} (Ours) & 64.706&77.647 \\
\hline
Routing Transformer & 83.529&75.294\\
Routing-\textit{HLogformer} (Ours) & 90.588&78.824 \\
\hline
Sparse Transformer & 69.412&  35.294 \\
Sparse-\textit{HLogformer} (Ours) & 72.941&38.823 \\
\bottomrule
\end{tabular}
}

\label{table:supervised_td}
\vspace{-0.15in}
\end{wraptable}

In addition to self-supervised learning, we also perform experiments on a supervised classification task. We utilize the \href{https://github.com/adanalvarez/TrailDiscover/tree/main}{TrailDiscover} dataset which contains two features for each data point: \texttt{"usedInWild"} which is a binary feature and takes two values of True or False, and \texttt{"MITRE Attack Tactics"} which is a feature that takes ten different values of attack type. 
We conduct the following two experiments: (1) a binary classification task on \texttt{"usedInWild"}, and (2) a multi-class classification task on \texttt{"MITRE Attack Tactics"}, and report average accuracy metric on each task.
The experimental results in Table~\ref{table:supervised_td}  show the significant improvement of our \textit{HLogformer} over the backbone transformers.

\subsection{Synthetic Anomaly Detection}

After conducting the self-supervised training, we obtain the representation of log data as well as the summary vector. Since we assume the model is trained with real data, fake data is likely to exhibit a different distribution or representation pattern compared to real data. Motivated by this, we can utilize the representations and summary vectors to perform fake data detection. To construct the fake dataset, we mismatch the key-value pairs in the real data with a probability of 
$
p=
0.2$. In this section, we divide the fake detection into three parts: (1) detection by loss, (2) detection by fake rate, and (3) detection by visualization.

\begin{wraptable}{r}{0.6\textwidth}
\centering
\vspace{-0.15in}
\caption{Synthetic anomaly detection by MLM/VHM loss. }
\begin{tabular}{c|c|c|c}
\toprule
\textbf{Data }& \textbf{CloudTrail} & \textbf{OKTA } &\textbf{TrailDiscover}\\
\hline
Real&3.925/0.575&3.379/0.578&4.580/1.833
\\
Fake&5.841/1.540&4.076/1.195&5.487/2.738\\
\bottomrule
\end{tabular}
\vspace{-0.2in}
\label{table:detect_loss}
\end{wraptable}
\textbf{Detection by loss.}
In this experiment, we train the model with the total loss as $\mathcal{L}_{MLM} +0.1\cdot\mathcal{L}_{VHM}$.
 As we train with the real data, we expect the MLM and VHM losses for real and fake data to show significant differences. Our results in Table~\ref{table:detect_loss}
 demonstrate that the losses for fake data are significantly higher than those for real data. This indicates that self-supervised learning effectively captures the hierarchical context information of real data.

\begin{wraptable}{r}{0.6\textwidth}
    \centering
    \vspace{-0.1in}
    \begin{small}
    \begin{tabular}{l|c|c|c|c|c}
        \toprule
        \textbf{Data} & \textbf{T=50} & \textbf{T=20} & \textbf{T=10} & \textbf{T=5} & \textbf{T=1} \\
        \hline
        Real& 4.08\% & 11.27\% & 18.03\% & 31.23\% & 73.54\% \\
        \hline
        Fake& 32.72\% & 45.36\% & 55.94\% & 68.73\% & 86.65\% \\
        \bottomrule
    \end{tabular}
    \end{small}
    \caption{Fake rate of different datasets under various $T$.}
    \vspace{-0.2in}
    \label{tab:fake_rate_diff_T}
\end{wraptable}

\textbf{Detection by fake rate.}
For each masked token $i$, we obtain an output probability $\hat{y}_{mask_i}$. We then construct a candidate set $\text{Candidate}_i$ with the top $T$ highest likelihoods. If the real value $x_{mask_i} \in \text{Candidate}_i$, we consider token $i$ as normal; otherwise, it is considered fake. Therefore, the fake rate can be calculated as:
$\text{Fake Rate} = \frac{ \text{number of fake tokens}}{ \text{number of all masked tokens}} \times 100\%.$
Table~\ref{tab:fake_rate_diff_T} shows the significant differences in fake rates across different datasets under various $T$ values.

\begin{wraptable}{r}{0.6\textwidth}
    \centering
    \vspace{-0.15in}
    \caption{Average accuracy at different threshold levels.}
    \begin{small}
    \begin{tabular}{l|c|c|c|c}
        \toprule
         \textbf{Threshold} $\alpha$  & $ 25\%$ & $  30\%$ & $ 35\%$ & $ 40\%$  \\
        \hline
        \textbf{Accuracy} &  91.92\% & 95.45\% & 95.96\% & 92.42\%  \\
        \bottomrule
    \end{tabular}  
    \end{small}
    \vspace{-0.15in}
    \label{tab:accuracy_diff_alpha}
\end{wraptable}
By setting different thresholds $\alpha$, we can predict whether a log is fake or not, i.e., a fake rate $> \alpha$ indicates the log is fake. 
Consequently, we can calculate the accuracy for both real and fake logs separately and then compute their average to determine the overall accuracy of the model.
With $T = 10$, we show the average accuracy at different threshold levels in Table~\ref{tab:accuracy_diff_alpha}. The results demonstrate that using the fake rate can achieve high accuracy in synthetic anomaly detection.

\begin{figure}[h!]
    \centering
\includegraphics[width=0.99\linewidth]{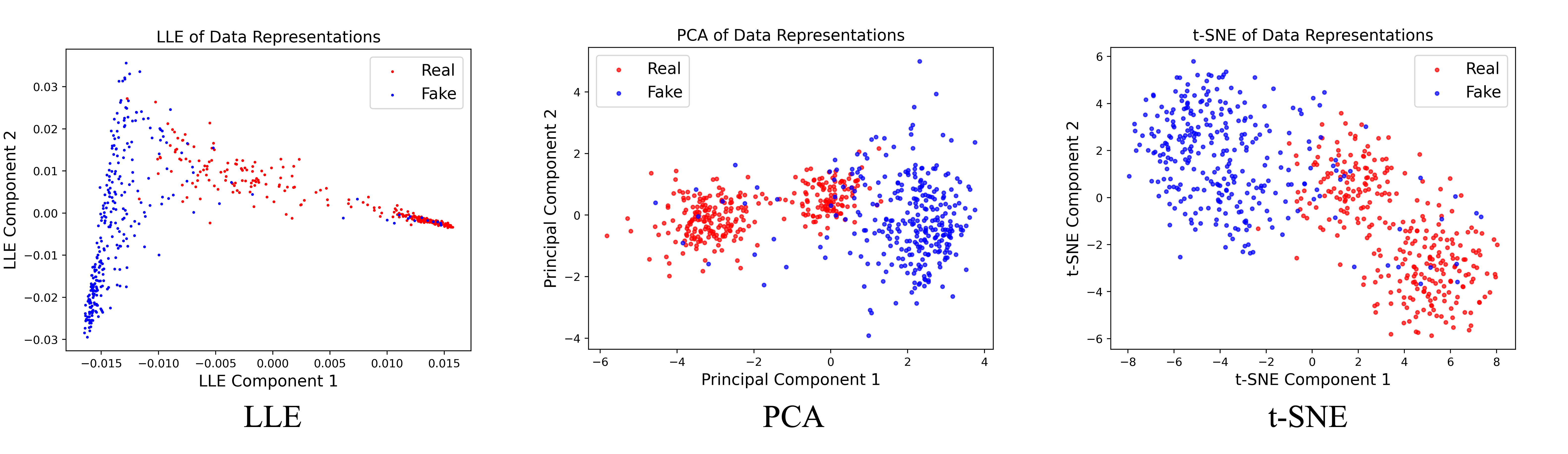}

    \caption{Visualization of summary vectors.}
    \label{fig:visual_ct}
\end{figure}

\textbf{Detection by visualization.} 
With VHM loss, we expect the summary vector of real data to be closely mapped to the center of the hypersphere. Consequently, the representations of real and fake data should exhibit significantly different patterns. To validate this, we use locally linear embedding (LLE)~\citep{roweis2000nonlinear}, principal component analysis (PCA) and t-distributed stochastic neighbor embedding (t-SNE)~\citep{van2008visualizing} to perform the dimension reduction and  visualization for the summary vectors obtained from real and fake data. The following figures in Figure~\ref{fig:visual_ct} demonstrate that the representations of these two data sources are separable and form distinct clusters. The visualization results on OKTA and TrailDiscover  can be found in the Figure~\ref{fig:visual_okta} and Figure~\ref{fig:visual_td}  in Appendix.

\subsection{Product Recommendation Task} 
\begin{wraptable}{r}{0.6\textwidth}
\vspace{-0.12in}
    \centering
    \begin{small}
    \begin{tabular}{c|c|c|c|c|c}
        \toprule
         \textbf{Precision-$K$ (\%)}  & 1&3&5&8&10  \\
        \hline
    \textbf{Transformer}&83.50&	80.67	&76.10&	70.81	&66.65\\
    \textbf{\textit{HLogformer}} &94.50&	89.17&81.90&	74.37&	69.45\\ 
        \bottomrule
    \end{tabular}  
    \end{small}
    \caption{Average precision at different $K$.}
    \label{tab:precision_recom}
    \vspace{-0.2in}
\end{wraptable}
To further demonstrate the effectiveness and advantages of our proposed \textit{HLogformer}, we conduct a product recommendation downstream task using the pretrained representations on Amazon Reviews dataset~\citep{hou2024bridging}. Specifically, we select 200 users with the highest number of purchased items and collect pretrained embeddings for all these items. For each user, the last 10 items purchased are treated as positive samples, while 10 items randomly selected from the available item repository are treated as negative samples. The average embedding of the remaining items is computed to represent the user's embedding. We then calculate the cosine similarity between each item's embedding (both positive and negative) and the user's embedding to generate a score list. This score list is sorted, and precision at \( K \) (precision-\( K \)) is computed based on the top \( K \) scores. Finally, we report the average precision-\( K \) across all users, as shown in Table~\ref{tab:precision_recom}. The results demonstrate a significant and consistent advantage of our \textit{HLogformer} over the vanilla transformer across all the $K$.


\subsection{Ablation Studies}

\begin{wraptable}{r}{0.6\textwidth}
\vspace{-0.15in}
\centering
\begin{small}
\begin{tabular}{l|c|c|c}
\toprule
\textbf{Architecture} & \textbf{CloudTrail }& \textbf{OKTA}&\textbf{TrailDiscover}  \\
\hline

\textit{HLogformer} &3.850&2.611&4.995\\

 w/o pretrained& 4.158&2.888&4.921 \\

w/o hierarchy  & 5.692&3.269&5.676  \\
w/o bi-direction&4.857&4.221&5.194\\
w/o summary &4.388&3.081&5.131\\
\bottomrule
\end{tabular}
\end{small}
\caption{Ablation studies. }
\label{table:ablation}
\vspace{-0.2in}
\end{wraptable}
To evaluate the contribution of each component in \textit{HLogformer}, we conduct ablation studies on three security log datasets. As shown in Table~\ref{table:ablation}, removing the hierarchical structure substantially increases the MLM loss, highlighting the importance of preserving the nested organization of log data. Disabling bidirectional propagation also degrades performance, indicating that information exchange across hierarchical levels is beneficial. Similarly, removing the summary vectors results in higher losses, confirming their effectiveness in transferring contextual information between segments. Pretrained initialization improves performance on CloudTrail and OKTA, although it provides no improvement on TrailDiscover, possibly due to a distribution mismatch. Overall, the hierarchy, bidirectional propagation, and summary vectors are key contributors to the performance of \textit{HLogformer}.

%% file: section/conclusion.tex
\section{Conclusion}

In this paper, we propose a novel and efficient hierarchical log transformer for dictionary-like log data. Our hierarchical transformers are specifically designed for log data such as CloudTrail and employ an adaptively recursive architecture tailored to this data. 
Our hierarchical framework is universal, making it orthogonal and compatible with various transformer backbones to further enhance performance and efficiency. Furthermore, our preliminary experiments show that the hierarchical representation learned through self-supervised learning exhibits great potential for encoding log data from events to groups and for various downstream tasks.

%% file: section/app.tex
\newpage

\input{section/related}

\newpage
\section{Supplemental Experiments}

\subsection{Summary vector visualization}
\begin{figure}[ht!]
    \centering
    \subfigure[OKTA.]{
        \includegraphics[width=0.48\linewidth]{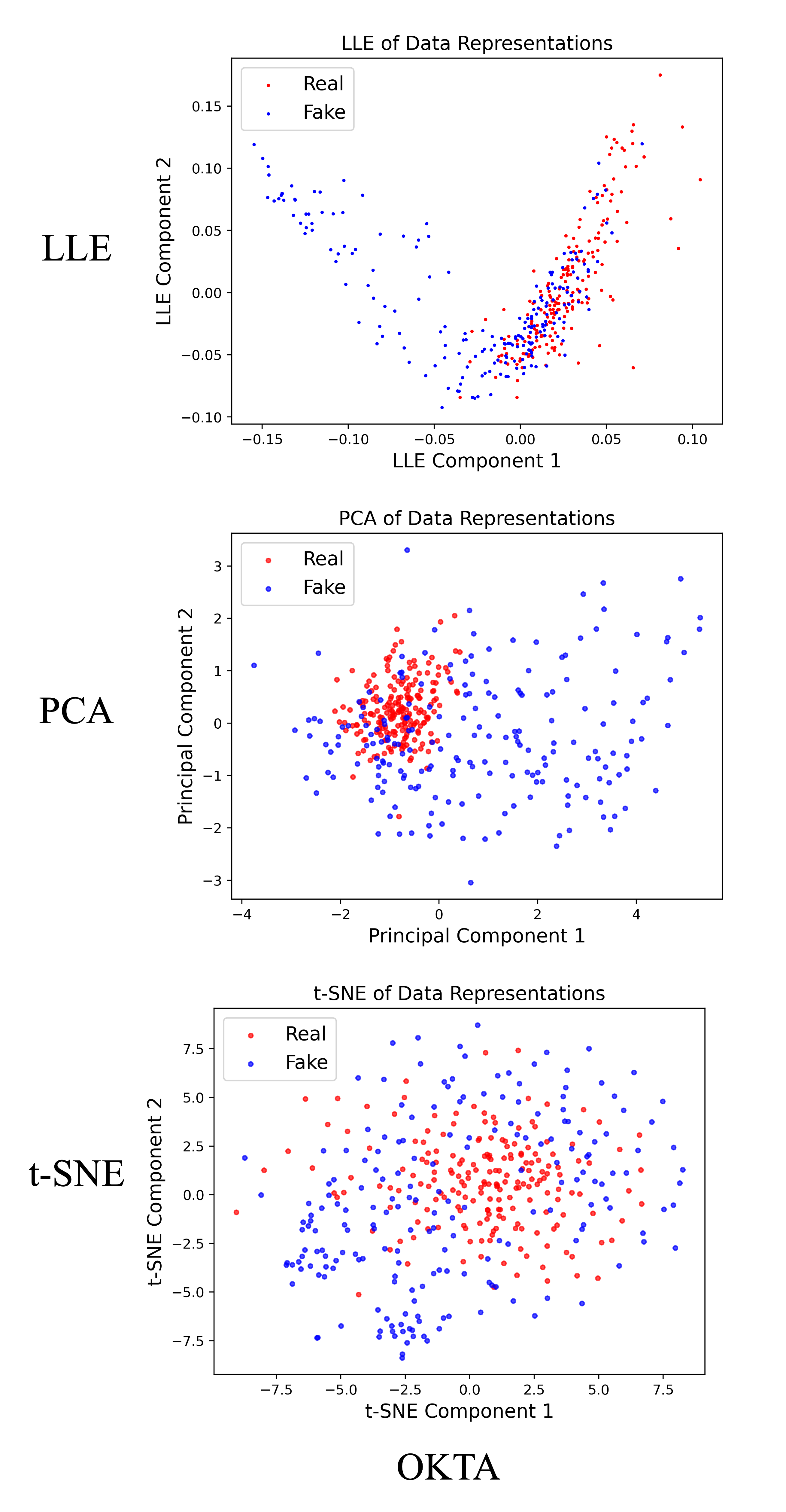}
        \label{fig:visual_okta}
    }
    \hfill
    \subfigure[TrailDiscover.]{
        \includegraphics[width=0.48\linewidth]{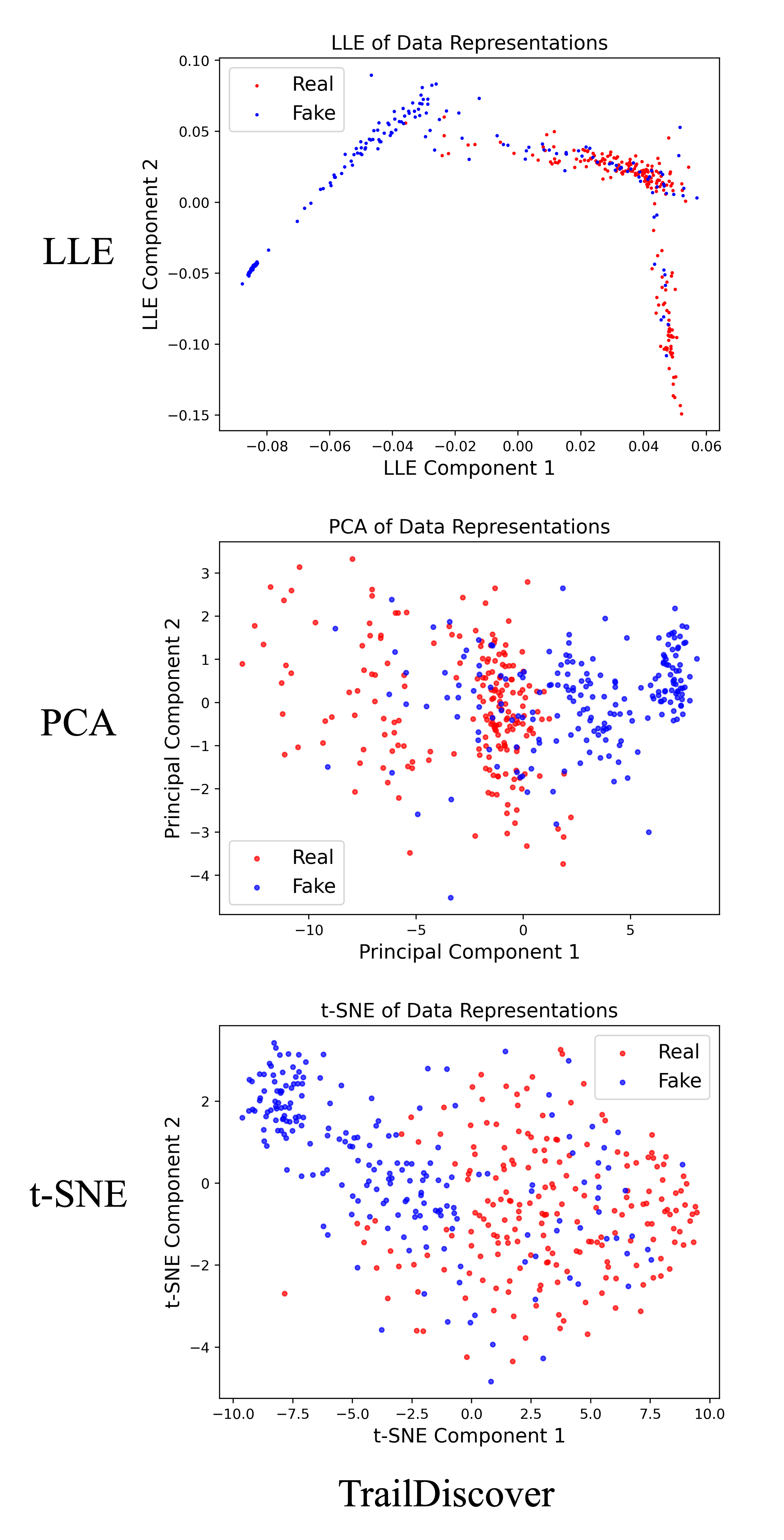}
        \label{fig:visual_td}
    }
    \caption{Visualization of summary vectors on different datasets.}
    \label{fig:visualization}
\end{figure}

With VHM loss, we expect the summary vector representations for the real and fake data to exhibit significantly different patterns. We visualize the learned summary vector representations using LLE, PCA and t-SNE on OKTA and TrailDiscover in Figure~\ref{fig:visual_okta} and Figure~\ref{fig:visual_td}. The results show evident separable clusters for real/fake data.

\newpage
\subsection{Training Loss Curve}
\begin{figure}[h!]
    \centering
    \includegraphics[width=0.5\linewidth]{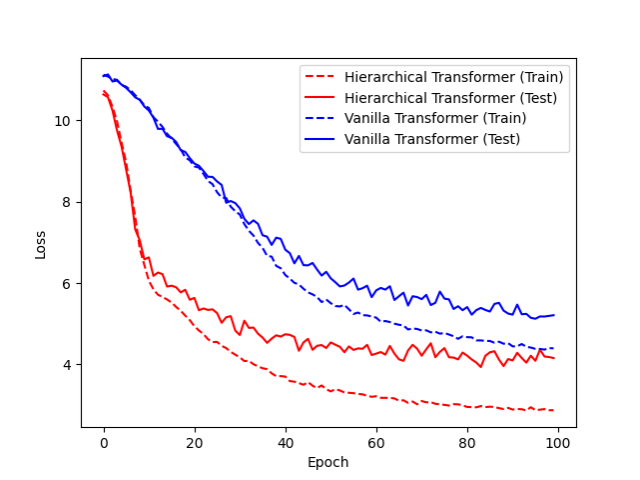}
    \caption{Loss curve during training.}
    \label{fig:training_loss}
\end{figure}

To validate the effectiveness of our hierarchical framework,
we track the training/testing loss during the training in Figure~\ref{fig:training_loss}. As can be observed in the curve, our hierarchical transformer exhibits faster convergence and better performance.

%% file: section/related.tex
\section{Related Works}
The related work in this area can be categorized into 2 main groups: efficient transformers including
incorporating global memory tokens, sparse attention mechanisms, segment-based recurrence methods, and hierarchical architectures. Each category offers distinct approaches to addressing the challenges of processing long sequences with transformers.
\subsection{Efficient Transformers}

\textbf{Global Memory Tokens in Transformers.}
Models like  Longformer~\citep{beltagy2020longformer}, ETC (Extended Transformer Construction)~\citep{ainslie2020etc}, and Big Bird~\citep{zaheer2020big} introduce global memory tokens to address the limitations of traditional transformers with long sequences. These tokens maintain attention connections to all other tokens in the sequence, allowing the models to capture broader contextual understanding while avoiding the quadratic memory and computational overhead of standard self-attention mechanisms. 

\noindent\textbf{Sparse Attention Mechanisms.}
Sparse transformers~\citep{child2019generating} employ fixed patterns with local and strided attention to address the inefficiencies of traditional transformers in processing long sequences. 
Other methods, such as those proposed by~\citep{roy2021efficient} and ~\citep{kitaev2020reformer}, enhance this concept by making the sparsity pattern learnable. These approaches adapt the attention patterns during training to better capture the data structure.
%
%

\noindent\textbf{Segment-based Recurrence.}
Segment-based recurrence methods, such as Transformer-XL~\citep{dai2019transformer} and Compressive Transformer~\citep{rae2019compressive}, introduce mechanisms to maintain and leverage contextual information across segments, significantly reducing memory and computational costs.

Despite their effectiveness, these approaches are not specifically tailored to the unique characteristics of log data, which often exhibit a hierarchical, dictionary-like structure. This gap underscores the need for models designed to capture and leverage the intrinsic structure of log data.

\subsection{Hierarchical Architectures}

Existing hierarchical transformer architectures~\citep{nawrot2021hierarchical, pappagari2019hierarchical, pan2021scalable, liu2021swin} that primarily focus on compressing or encoding fine-grained information and decoding it back to the original size if necessary.  For example, Hourglass~\citep{nawrot2021hierarchical} utilizes downsampling and upsampling techniques to create hierarchical and efficient transformers.
\citet{pappagari2019hierarchical} design hierarchical transformers by segmenting the input into smaller chunks and feeding each chunk into the base model, effectively managing long documents.
Swin Transformer~\citep{liu2021swin} employs a shifted windows scheme to design an efficient hierarchical architecture.
sentence-level information in text data.
However, these architectures often prioritize compression and encoding efficiency over accurately representing the hierarchical nature of data. They focus on reducing the size of the data for efficient processing and storage, and then decoding it back when needed. These approaches do not fully align with the unique characteristics of log data, which require capturing and leveraging their inherent hierarchical structure.

%% file: aaai25.bib
@article{roweis2000nonlinear,
  title={Nonlinear dimensionality reduction by locally linear embedding},
  author={Roweis, Sam T and Saul, Lawrence K},
  journal={science},
  volume={290},
  number={5500},
  pages={2323--2326},
  year={2000},
  publisher={American Association for the Advancement of Science}
}

@article{child2019generating,
  title={Generating long sequences with sparse transformers},
  author={Child, Rewon and Gray, Scott and Radford, Alec and Sutskever, Ilya},
  journal={arXiv preprint arXiv:1904.10509},
  year={2019}
}

@article{beltagy2020longformer,
  title={Longformer: The long-document transformer},
  author={Beltagy, Iz and Peters, Matthew E and Cohan, Arman},
  journal={arXiv preprint arXiv:2004.05150},
  year={2020}
}

@article{ainslie2020etc,
  title={ETC: Encoding long and structured inputs in transformers},
  author={Ainslie, Joshua and Ontanon, Santiago and Alberti, Chris and Cvicek, Vaclav and Fisher, Zachary and Pham, Philip and Ravula, Anirudh and Sanghai, Sumit and Wang, Qifan and Yang, Li},
  journal={arXiv preprint arXiv:2004.08483},
  year={2020}
}

@article{zaheer2020big,
  title={Big bird: Transformers for longer sequences},
  author={Zaheer, Manzil and Guruganesh, Guru and Dubey, Kumar Avinava and Ainslie, Joshua and Alberti, Chris and Ontanon, Santiago and Pham, Philip and Ravula, Anirudh and Wang, Qifan and Yang, Li and others},
  journal={Advances in neural information processing systems},
  volume={33},
  pages={17283--17297},
  year={2020}
}

@article{roy2021efficient,
  title={Efficient content-based sparse attention with routing transformers},
  author={Roy, Aurko and Saffar, Mohammad and Vaswani, Ashish and Grangier, David},
  journal={Transactions of the Association for Computational Linguistics},
  volume={9},
  pages={53--68},
  year={2021},
  publisher={MIT Press One Rogers Street, Cambridge, MA 02142-1209, USA journals-info~…}
}

@article{kitaev2020reformer,
  title={Reformer: The efficient transformer},
  author={Kitaev, Nikita and Kaiser, {\L}ukasz and Levskaya, Anselm},
  journal={arXiv preprint arXiv:2001.04451},
  year={2020}
}

@inproceedings{katharopoulos2020transformers,
  title={Transformers are rnns: Fast autoregressive transformers with linear attention},
  author={Katharopoulos, Angelos and Vyas, Apoorv and Pappas, Nikolaos and Fleuret, Fran{\c{c}}ois},
  booktitle={International conference on machine learning},
  pages={5156--5165},
  year={2020},
  organization={PMLR}
}

@article{dai2019transformer,
  title={Transformer-xl: Attentive language models beyond a fixed-length context},
  author={Dai, Zihang and Yang, Zhilin and Yang, Yiming and Carbonell, Jaime and Le, Quoc V and Salakhutdinov, Ruslan},
  journal={arXiv preprint arXiv:1901.02860},
  year={2019}
}

@article{rae2019compressive,
  title={Compressive transformers for long-range sequence modelling},
  author={Rae, Jack W and Potapenko, Anna and Jayakumar, Siddhant M and Lillicrap, Timothy P},
  journal={arXiv preprint arXiv:1911.05507},
  year={2019}
}

@article{chevalier2023adapting,
  title={Adapting language models to compress contexts},
  author={Chevalier, Alexis and Wettig, Alexander and Ajith, Anirudh and Chen, Danqi},
  journal={arXiv preprint arXiv:2305.14788},
  year={2023}
}

@article{vaswani2017attention,
  title={Attention is all you need},
  author={Vaswani, Ashish and Shazeer, Noam and Parmar, Niki and Uszkoreit, Jakob and Jones, Llion and Gomez, Aidan N and Kaiser, {\L}ukasz and Polosukhin, Illia},
  journal={Advances in neural information processing systems},
  volume={30},
  year={2017}
}

@inproceedings{ruff2018deep,
  title={Deep one-class classification},
  author={Ruff, Lukas and Vandermeulen, Robert and Goernitz, Nico and Deecke, Lucas and Siddiqui, Shoaib Ahmed and Binder, Alexander and M{\"u}ller, Emmanuel and Kloft, Marius},
  booktitle={International conference on machine learning},
  pages={4393--4402},
  year={2018},
  organization={PMLR}
}

@article{dosovitskiy2020image,
  title={An image is worth 16x16 words: Transformers for image recognition at scale},
  author={Dosovitskiy, Alexey and Beyer, Lucas and Kolesnikov, Alexander and Weissenborn, Dirk and Zhai, Xiaohua and Unterthiner, Thomas and Dehghani, Mostafa and Minderer, Matthias and Heigold, Georg and Gelly, Sylvain and others},
  journal={arXiv preprint arXiv:2010.11929},
  year={2020}
}

@article{velivckovic2017graph,
  title={Graph attention networks},
  author={Veli{\v{c}}kovi{\'c}, Petar and Cucurull, Guillem and Casanova, Arantxa and Romero, Adriana and Lio, Pietro and Bengio, Yoshua},
  journal={arXiv preprint arXiv:1710.10903},
  year={2017}
}

@article{nawrot2021hierarchical,
  title={Hierarchical transformers are more efficient language models},
  author={Nawrot, Piotr and Tworkowski, Szymon and Tyrolski, Micha{\l} and Kaiser, {\L}ukasz and Wu, Yuhuai and Szegedy, Christian and Michalewski, Henryk},
  journal={arXiv preprint arXiv:2110.13711},
  year={2021}
}

@article{van2008visualizing,
  title={Visualizing data using t-SNE.},
  author={Van der Maaten, Laurens and Hinton, Geoffrey},
  journal={Journal of machine learning research},
  volume={9},
  number={11},
  year={2008}
}

@article{devlin2018bert,
  title={Bert: Pre-training of deep bidirectional transformers for language understanding},
  author={Devlin, Jacob and Chang, Ming-Wei and Lee, Kenton and Toutanova, Kristina},
  journal={arXiv preprint arXiv:1810.04805},
  year={2018}
}

@inproceedings{hou2024protransformer,
  title={ProTransformer: Robustify Transformers via Plug-and-Play Paradigm},
  author={Hou, Zhichao and Gao, Weizhi and Shen, Yuchen and Liu, Xiaorui},
  booktitle={ICLR 2024 Workshop on Reliable and Responsible Foundation Models},
  year={2024}
}

@article{hou2024bridging,
  title={Bridging language and items for retrieval and recommendation},
  author={Hou, Yupeng and Li, Jiacheng and He, Zhankui and Yan, An and Chen, Xiusi and McAuley, Julian},
  journal={arXiv preprint arXiv:2403.03952},
  year={2024}
}

@inproceedings{pappagari2019hierarchical,
  title={Hierarchical transformers for long document classification},
  author={Pappagari, Raghavendra and Zelasko, Piotr and Villalba, Jes{\'u}s and Carmiel, Yishay and Dehak, Najim},
  booktitle={2019 IEEE automatic speech recognition and understanding workshop (ASRU)},
  pages={838--844},
  year={2019},
  organization={IEEE}
}

@inproceedings{liu2021swin,
  title={Swin transformer: Hierarchical vision transformer using shifted windows},
  author={Liu, Ze and Lin, Yutong and Cao, Yue and Hu, Han and Wei, Yixuan and Zhang, Zheng and Lin, Stephen and Guo, Baining},
  booktitle={Proceedings of the IEEE/CVF international conference on computer vision},
  pages={10012--10022},
  year={2021}
}

@inproceedings{pan2021scalable,
  title={Scalable vision transformers with hierarchical pooling},
  author={Pan, Zizheng and Zhuang, Bohan and Liu, Jing and He, Haoyu and Cai, Jianfei},
  booktitle={Proceedings of the IEEE/cvf international conference on computer vision},
  pages={377--386},
  year={2021}
}

@article{huang2020tabtransformer,
  title={Tabtransformer: Tabular data modeling using contextual embeddings},
  author={Huang, Xin and Khetan, Ashish and Cvitkovic, Milan and Karnin, Zohar},
  journal={arXiv preprint arXiv:2012.06678},
  year={2020}
}

@article{wu2024equivariant,
  title={Equivariant spatio-temporal attentive graph networks to simulate physical dynamics},
  author={Wu, Liming and Hou, Zhichao and Yuan, Jirui and Rong, Yu and Huang, Wenbing},
  journal={Advances in Neural Information Processing Systems},
  volume={36},
  year={2024}
}

@inproceedings{hou2025robustness,
  title={Robustness reprogramming for representation learning},
  author={Hou, Zhichao and Torkamani, MohamadAli and Krim, Hamid and Liu, Xiaorui},
  booktitle={International Conference on Learning Representations},
  volume={2025},
  pages={53682--53704},
  year={2025}
}

@article{hou2026tefm,
  title={TEFM: Token-Efficient Faithful Modeling for Structured Data},
  author={Hou, Zhichao and Sha, Lingdao and Mao, Xueyu and Liu, Yang and Qiu, Peijie and Song, Rui},
  journal={arXiv preprint arXiv:2609.09552},
  year={2026}
}

@article{zhang2026centile,
  title={CENTILE: A Telemetry Foundation Model Evaluated by the Decisions It Drives},
  author={Zhang, Zifan and Hou, Zhichao and Ji, Tingxiang and Liu, Yuchen},
  journal={arXiv preprint arXiv:2608.01725},
  year={2026}
}
